# LOW-LEVEL FEATURES FOR IMAGE RETRIEVAL BASED ON EXTRACTION OF DIRECTIONAL BINARY PATTERNS AND ITS ORIENTED GRADIENTS HISTOGRAM

Nagaraja S. and Prabhakar C.J.

Department of P.G. Studies and Research in Computer Science
Kuvempu University, India

*ABSTRACT*

*In this paper, we present a novel approach for image retrieval based on extraction of low level features using techniques such as Directional Binary Code (DBC), Haar Wavelet transform and Histogram of Oriented Gradients (HOG). The DBC texture descriptor captures the spatial relationship between any pair of neighbourhood pixels in a local region along a given direction, while Local Binary Patterns (LBP) descriptor considers the relationship between a given pixel and its surrounding neighbours. Therefore, DBC captures more spatial information than LBP and its variants, also it can extract more edge information than LBP. Hence, we employ DBC technique in order to extract grey level texture features (texture map) from each RGB channels individually and computed texture maps are further combined which represents colour texture features (colour texture map) of an image. Then, we decomposed the extracted colour texture map and original image using Haar wavelet transform. Finally, we encode the shape and local features of wavelet transformed images using Histogram of Oriented Gradients (HOG) for content based image retrieval. The performance of proposed method is compared with existing methods on two databases such as Wang's corel image and Caltech 256. The evaluation results show that our approach outperforms the existing methods for image retrieval.*

## 1. INTRODUCTION

In last two decades, exponentially increase of digital images due to increase in availability of digital cameras, mobile devices and other hand held devices. These things with Internet have created a way to generate and publish visual content in web. Recently, the number of Internet users is growing in this way, visual information also increasing in the web, which becomes the largest and most heterogeneous digital image database. A new research challenges are transferring, processing, archiving and retrieving of these digital image databases. Among all the challenges involved in digital image databases, the retrieving the images from databases is an important and most attractive research topic for computer vision community due to its applications and various research problems. The image retrieval from the image database is the application of computer vision techniques for searching the digital images in large databases. Generally three types of methods are used for image retrieval i.e. semantic-based, text-based and content-based techniques. Normally, web-based search engines are used for retrieval of images from image database based on text-based technique using text as a keyword. The survey on text-based and semantic-based image retrieval approaches can be found in [31], [32]. Text-based image retrieval systems use traditional database methods to manage images and human intervention is more, because every image in the database is indexed properly before retrieving the images. Having humans manually make notes on images by entering keywords in a huge database can be time consuming and may not capture the keywords desired to describe the image. In semantic-based image retrieval, the user has to specify the query through a natural language description based on visual concepts of interest and the image in the image database are annotated with semantic keywords. Content-based image retrieval (CBIR) methods use visual contents to search images from the large digital image databases has been an active research area in the computer vision and pattern recognition.

Content-based means that the retrieving the images based on analyzing the contents of the image rather than the keywords, tags, or descriptions associated with the image. The content of the image that is widely used are colours, shapes, textures, or any other information of the image itself. Y. Liu et.al [1] conducted survey on content-based image retrieval techniques and they covered steps involved in various CBIR approaches such as low-level feature extraction from the image, similarity measurement and high-level semantic features. The technique which uses other information of the images such as depth is proposed by A. Jain et al. [11] for retrieving images from the large database based on engineering/ computer-aided design (CAD) models. They proposed a linear approximation procedure that can capture the depth information based on shape from shading, then, similarity measure combines shape and depth information to retrieval of images from the database. Another category of techniques which uses bag-of-visual features apart from the low level features. J. Yu et al. [10] investigate various combinations of mid-level features to build image retrieval system using the bag-of-features model.

Initially, CBIR techniques were developed based on any one of the features such as colour and texture alone. One such feature, which was used alone for CBIR was colour. Colour is one of the important features in the field of content-based image retrieval. Colour histogram [14] based image retrieval is simple to implement and had been well used in CBIR system. The histogram reflects the statistical distribution of the intensities of the three colour channels. Colour histogram is computed by discretizing the colours within the image and counting the number of pixels of each colour. Many authors [15], [16] have proposed colour descriptors to exploit special information, together with compact colour central moments and colour coherence vector. The retrieval performance of these colour descriptors is limited due to insufficiency in discrimination power compared to other features such as texture and shape.

It is found that usage of single type feature is not sufficient in order to achieve high retrieval rate. Hence, the researchers have focused on investigating techniques based on combination of low-level features such as colour, texture and shape. The survey on recent techniques which uses combination of low-level features is conducted by O.A.B. Penatti et al. [6] and they compared various global colour and texture descriptors for web image retrieval. Their analysis of the correlation is provided for the best descriptors, which provides hints at the end best opportunities of their use in combination. P.S. Hiremath et al. [2] proposed a technique, which combines low-level features such as colour, texture and shape for content based image retrieval. In this method, initially the image is partitioned into non-overlapping tiles of equal size then extracted colour and texture features from Gabor filter responses and shape information is obtained from Gradient Vector Flow fields. Most Similar Highest Priority (MSHP) principle and adjacency matrix of a bipartite graph is used to match the query and target image. B. Tao et al. [4] presented a method for texture recognition and image retrieval using gradient indexing. Local activity spectrum is used for image retrieval and sum of minimum distance is employed for matching. S.M. Youssef [5] proposed Integrated Curvelet-based image retrieval method, which integrates curvelet multiscale ridgelets with region based vector codebook sub-band clustering for extraction of dominant colours and texture. K. Iqbal et al. [13] present a content-based image retrieval approach for biometric security based on colour, texture and shape features controlled by fuzzy heuristics.

V. Takala et al. [3] proposed two methods based on extraction of texture within blocks of image. Local Binary Pattern (LBP) is used to extract texture based feature from images. In the first method, they divide the query and database images into equally sized blocks then LBP histograms are extracted, which were matched using a relative $L_1$ dissimilarity measure based on the Minkowski distances. In the second method, image division on database images and calculated a single feature histogram for the query image and then find the best match by sliding search window. X. Yuan et al. [7] proposed a new CBIR technique based on bag-of-features model by integrating scale invariant feature transform (SIFT) and local binary pattern (LBP). The combination of SIFT and LBP are used to derived two methods: patch-based and image-based integration. These techniques yield complementary and substantial improvement on image retrieval even in the case of noisy background and ambiguous objects.

H. Sebai et al. [8] present an adaptive content based image retrieval method based on 3D-LBP and Histogram of Orientated Gradients (HOG) in order to extract colour, texture and shape feature of the image. The aim of the their proposed method is to increase the performance by optimizing image features selection according to image nature such as textured and structured and at the same time maintaining a small sized feature to attain better matching and lower the complexity. S.K. Vipparthi et al. [9] proposed colour directional local quinary patterns for content based indexing and retrieval. They proposed novel descriptor to extracts colour-texture features for image retrieval application called Color Directional Local Quinary Pattern (CDLQP). This descriptor extracts R, G and B channel wise directional edge information separately between reference pixel and its neighbourhoods by computing its grey-level difference based on quinary value instead of binary and ternary value in four directions on an image. The main drawback of CDLQP descriptor is that computation cost is very high due to quinary pattern compared to binary pattern of LBP. S. Banerji et al. [12] proposed novel image descriptors based on colour, texture, shape and wavelets for object and scene image classification. First, they introduced a new three dimensional local binary patterns (3D-LBP) descriptor for encoding colour and texture information of the image. The 3D-LBP descriptor produces three colour images from the original colour image then Haar wavelet transform is applied to the three 3D-LBP images and original colour image. Finally, Histograms of Oriented Gradients (HOG) is calculated from the Haar wavelet transformed images for encoding shape and local features and form a new H-descriptor.
Directional Binary Code (DBC) is proposed by B. Zhang et al. [21] for face texture analysis. The DBC captures the spatial relationship between any pair of neighbourhood pixels in a local region along a given direction, while LBP variants such as 3D-LBP and CDLQP consider the relationship between a given pixel and its surrounding neighbours. Therefore, DBC captures more spatial information and edge information than LBP variants. The advantages of DBC motivated us to propose the novel CBIR technique based on colour texture features extracted using DBC. The Fig. 1 shows the block diagram of proposed method. Our technique combines three techniques such as DBC, Haar Wavelet and HOG employed sequentially. We employ DBC technique in order to extract grey level texture features from each RGB channel individually and computed texture maps are combined which represent colour texture features of an image. Then, we decomposed the extracted colour texture map and original image using Haar wavelet transform. Finally, we encode the shape and local features of wavelet transformed images using Histogram of Oriented Gradients (HOG). These sequences of steps are employed on training images and created the feature vectors, in offline mode. The feature vector of the query image is obtaining by following the same sequence of steps and finally similarity measure is employed to retrieve the images from database.

The remaining section of the paper is organized as follows. The section 2 describes theoretical background for DBC. The Haar Wavelet transform is explained in section 3. The HOG descriptor and its sequence of steps are explained in the section 4. The experimental results are illustrated in the section 5. Finally, the paper is concluded.

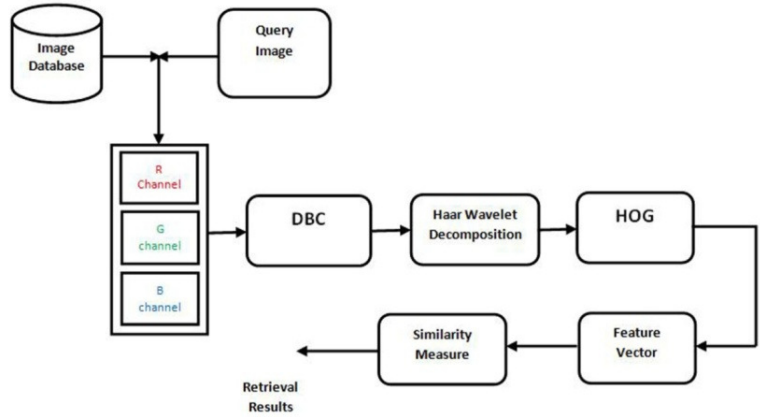

Figure 1. Our proposed image retrieval method

## 2. EXTRACTION OF COLOUR TEXTURE FEATURES USING DBC

Texture is one of the most important features in an image. Texture analysis has been widely used in CBIR systems due to its potential value. Textures may be described according to their spatial, frequency or perceptual properties. Different algorithms are proposed for texture analysis such as, Wavelet Correlogram [17], Gabor Wavelet Correlogram [18], Gray Level Coocurrence Matrices [19], Local Binary Patterns (LBP) [20] and its variants. LBP is computationally simple, efficient, good texture discriminative property and has been highly successful for various computer vision problems like image retrieval, face analysis, dynamic texture recognition, and motion analysis etc. LBP has a property that favours its usage in interest region description such as computationally simplicity and tolerance against illumination changes. LBP operator is structural and statistical texture descriptor in terms of the characteristics of the local structure, so that it is most powerful for texture analysis.

The original LBP was introduced by Ojala et al. [20] in 1996; it was mainly derived for texture analysis. The LBP operator labels the pixels of an image by using the 3x3 neighbourhood of each pixel with the center value as a threshold and the result as a binary number is defined as

$$LBP_{P,R} = \sum_{p=0}^{P-1} s(g_p - g_c) 2^p, \qquad (1)$$

$$s(x) = \begin{cases} 1 & x \geq 0 \\ 0 & x < 0 \end{cases}, \qquad (2)$$

where $P$ is the number of neighbourhoods, $R$ is the radius of the neighbourhood, $g_c$ is the grey value of the centre pixel and $g_p$ is the grey value of its neighbourhoods. LBP considers the relationship between a given pixel and its surrounding neighbours. But it won't capture spatial information and also it can't extract more edge information. Directional Binary Code (DBC) is proposed by B. Zhang et al. [21] captures the spatial relationship between any pair of neighbourhood pixels in a local region along a given direction, while LBP variants consider the relationship between a given pixel and its surrounding neighbours. Therefore, DBC captures more spatial information and edge information than LBP variants. The DBC is encoding the directional edge information in a neighbourhood of an image. In a given image $I$ calculate its first-order derivatives along $0^0$, $45^0$, $90^0$ and $135^0$ directions as $I'_{\alpha,d}$, where $\alpha = 0^0, 45^0, 90^0$ and $135^0$ and $d$ is the distance between the given pixel and its neighbouring pixel. Let $x_{i,j}$ be a point in $I$, then the four directional derivatives at $x_{i,j}$ are defined as

$$I'_{0^0,d}(x_{i,j}) = I(x_{i,j}) - I(x_{i,j-d})$$

$$I'_{45^0,d}(x_{i,j}) = I(x_{i,j}) - I(x_{i-d,j+d})$$

$$I'_{90^0,d}(x_{i,j}) = I(x_{i,j}) - I(x_{i-d,j})$$

$$I'_{135^0,d}(x_{i,j}) = I(x_{i,j}) - I(x_{i-d,j-d}) \quad (3)$$

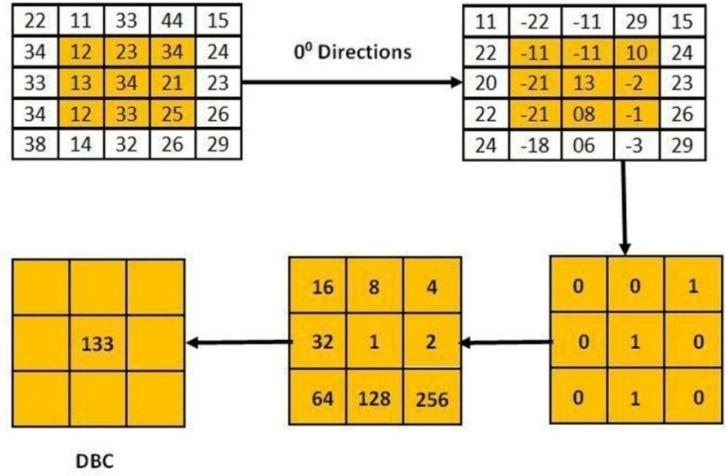

Figure 2. DBC pattern along $0^0$ directions.

A thresholding function $f\left(I'_{\alpha,d}(x)\right)$ is defined below and is applied to the four directional derivatives to output a binary code in the given direction.

$$f\left(I'_{\alpha,d}(x)\right) = \begin{cases} 1, & if\ I'_{\alpha,d}(x) \geq 0 \\ 0, & if\ I'_{\alpha,d}(x) < 0 \end{cases} \quad (4)$$

Using equation (2), the DBC ($\alpha = 0^0$ (shown in Fig. 2), $45^0$, $90^0$ and $135^0$) is defined as

$$DBC_{\alpha,d}(x_{i,j}) = \begin{Bmatrix} f\left(I'_{\alpha,d}(x_{i,j})\right); f\left(I'_{\alpha,d}(x_{i,j-d})\right); f\left(I'_{\alpha,d}(x_{i-d,j-d})\right), f\left(I'_{\alpha,d}(x_{i-d,j})\right); f\left(I'_{\alpha,d}(x_{i-d,j+d})\right); \\ f\left(I'_{\alpha,d}(x_{i,j+d})\right); f\left(I'_{\alpha,d}(x_{i+d,j+d})\right); f\left(I'_{\alpha,d}(x_{i+d,j})\right); f\left(I'_{\alpha,d}(x_{i+d,j-d})\right) \end{Bmatrix} \quad (5)$$

Colour is an essential attribute of an image and it provides more information than a single or grey value. There have been few attempts to incorporate chrominance information into textural features. A colour texture can be regarded as a pattern described by the relationship between its chromatic and structural distribution. The colour texture features are extracted from each channel of colour image using DBC technique, which yields grey level texture features for each colour band and further combined the computed grey level texture maps, which represents colour texture features of an image.

## 3. HAAR WAVELET TRANSFORM

We employ Haar wavelet transform [24] on colour texture map and its original colour image in order to extract the local information for enhancing local contrast. Another advantage of the Haar Wavelet transform is that it reduces dimensionality by preserving more texture as well as colour and shape information in the form of coefficients obtained by dividing the image into four sub-bands. Haar wavelet is used due to its simplicity and computational efficiency. The two dimensional Haar wavelet transform is defined as the projection of an image on to the two dimensional Haar basis functions, which are formed by the tensor product of the one dimensional Haar scaling and wavelet functions [24], [25]. The Haar scaling function $\phi(x)$ is defined as [26]

$$\phi(x) = \begin{cases} 1, & 0 \leq x < 1 \\ 0, & otherwise \end{cases} \quad (6)$$

A family of functions can be generated from the basic scaling function by scaling and translation

$$\phi_{i,j}(x) = 2^{\frac{i}{2}} \phi(2^i x - j) \quad (7)$$

As a result, the scaling functions $\phi_{i,j}(x)$ can span the vector spaces $V^i$, which are nested as follows

$$V^0 \subset V^1 \subset V^2 \subset \cdots \quad (8)$$

The Haar wavelet function $\psi(x)$ is defined as

$$\psi(x) = \begin{cases} 1, & 0 \leq x < 1/2 \\ -1, & 1/2 \leq x < 1 \\ 0, & otherwise \end{cases} \quad (9)$$

The Haar wavelets are generated from the mother wavelet by scaling and translation

$$\psi_{i,j}(x) = 2^{i/2} \psi(2^i x - j) \quad (10)$$

The Haar wavelets $\psi_{i,j}(x)$ span the vector space $W^i$, which is the orthogonal complement of $V^i$ in $V^{i+1}$:

$$V^{i+1} = V^i \oplus W^i \quad (11)$$

The two dimensional Haar basis functions are the tensor product of the one dimensional scaling and wavelet functions.

## 4. EXTRACTION OF SHAPE FEATURES USING HOG DESCRIPTOR

In order to extract the shape features, we employed Histograms of Oriented Gradients (HOG) technique [22][23] proposed by N. Dalal et al. [27] on each sub-band of wavelet transformed image, which stores the information about the shapes contained in the image, represented by histograms of the slopes of the object edges. Each bin in the histogram represents the number of edges that have orientations within a certain angular range. The concatenation of computed histograms of all the four sub-bands yields the HOG descriptor, which is stored the shape as well as texture information, which can be used for content based image retrieval. Since, DBC and Haar wavelet transform are employed in order to enhance the edges and other high-frequency local features, the choice of HOG yield more shape information of an image with enhanced edges than unprocessed image. The following steps are used to compute the local histograms of gradient. First step is to compute the gradients of the image, in second step the histograms of orientation is build for each cell and finally normalize the histograms within each block of cells.

*Gradient Computation*: The gradient of an image $I$ is obtained by filtering it with horizontal and vertical one dimensional discrete derivative mask.

$$D_X = [-1 \ 0 \ 1] \ \text{and} \ D_Y = \begin{bmatrix} 1 \\ 0 \\ -1 \end{bmatrix}, \qquad (12)$$

where $D_X$ and $D_Y$ are horizontal and vertical masks respectively and obtain the $X$ and $Y$ derivatives using following convolution operation.

$$I_X = I * D_X \quad \text{and} \quad I_Y = I * D_Y \qquad (13)$$

The magnitude of the gradient is

$$|G| = \sqrt{I_X^2 + I_Y^2} \qquad (14)$$

The orientation of the gradient is given by

$$\theta = arctan\frac{I_Y}{I_X} \qquad (14)$$

*Orientation Binning*: In the second step creating the cell histograms. Each pixel calculates a weighted vote for an orientation based histogram channel based on the values found in the gradient computation. The cells themselves are rectangular and the histogram channels are evenly spread over $0^0$ to $180^0$ or $0^0$ to $360^0$, depending on whether the gradient is unsigned or signed. N. Dalal and B. Triggs found that unsigned gradients used in conjunction with 9 histogram channels performed best in their experiments.

*Descriptor Blocks*: In order to changes in illumination and contrast, the gradient strengths should be regionally normalized, which needs grouping the cells together into larger spatially connected blocks. The HOG descriptor is then the vector of the elements of the normalized cell histograms from all of the block regions. These blocks generally overlap, that means each cell contributes more than once to the final descriptor.

A normalization factor is then computed over the block and all histograms within this block are normalized according to this normalization factor. Once this normalization step has been performed all the histograms will concatenated in a single feature vector. There are different methods for block normalization. Let $v$ be the non-normalized vector containing all histograms in a given block, $\|v_k\|$ be its $k$-norm for $k=1, 2$ and $e$ be some small constant. The normalization factor $f$ can obtain by these methods.

$$\text{L1-norm: } f = \frac{v}{\|v\|_1 + e} \qquad (15)$$

$$\text{L2-norm: } f = \frac{v}{\sqrt{\|v\|_2^2 + e^2}} \tag{16}$$

## 5. EXPERIMENTAL RESULTS

We conducted experiments in order to assess our method for image retrieval using two popular datasets, namely the Wang's Corel image dataset [29] and Caltech 256 dataset [30], intermediate results are shown in Fig. 3. The image retrieval system performance of any feature descriptor not only depends on feature extraction method, but also requires good similarity metrics. Various researchers used different methods to measure similarity to obtain relevant results. In our proposed method, we used Euclidean distance with minimum distance classifier. The four similarity distance measures which are used for comparison are defined below

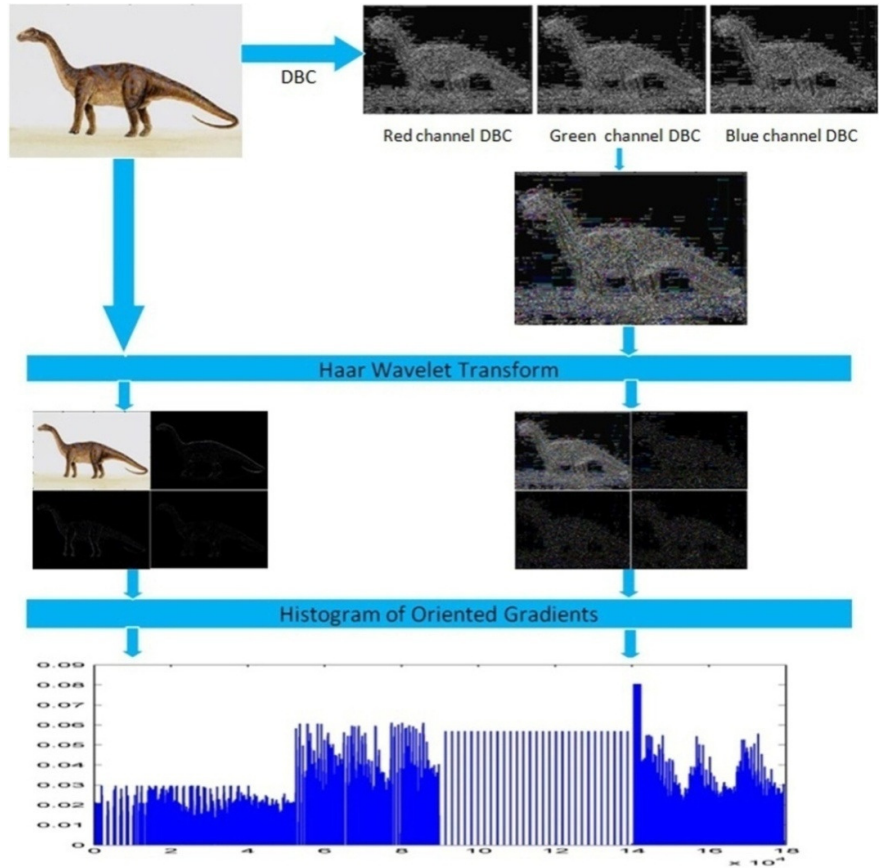

Figure 3. The intermediate results of our approach for sample image of the Wang's database

$$Manhattan\ or\ L_1\ city-block\ Distance: D(X^k, X^t) = \sum_{i=1}^{n} |X^k - X^t|, \tag{17}$$

$$Euclidean \ or \ L_2 \ Distance: D(X^k, X^t) = \sqrt{\sum_{i=1}^{n}(X_i^k - X_i^t)^2}, \qquad (18)$$

$$Candberra \ Distance: D(X^k, X^t) = \sum_{i=1}^{n}\frac{|X^k - X^t|}{|X^k + X^t|}, \qquad (19)$$

$$Chi-square \ Distance: D(X^k, X^t) = \sum_{i=1}^{n}\frac{(X^k - X^t)^2}{(X^k + X^t)}, \qquad (20)$$

where $X^k$ and $X^t$ are feature vectors of query and training image respectively and $n$ is the length of feature vector.

### 5.1. Experiment on Wang's Database

Wang's image dataset [29] contains 1000 Corel images with ground truth. These images are grouped into 10 categories with each category containing 100 images of the size 256x384 or 384x256. The images in the same group are considered as similar images. The images are subdivided into 10 classes such that it is almost sure that a user wants to find the other images from a class if the query is from one of these 10 classes. This database was used extensively to test the different features because the size of the database and the availability of class information. The Fig. 4 shows the sample Corel images of Wang's database. The proposed method is compared with other methods like patch based SIFT-LBP [7], image based SIFT-LBP [7], Histogram based [28] and colour, shape and texture based method [2].

The performance of our image retrieval method can be measured in terms of precision and recall. Precision measures the ability of the system to retrieve only relevant models, while recall measure the ability of the system to retrieve all relevant models and are defined as

$$P = \frac{Number \ of \ relevant \ images \ retrieved}{Total \ number \ of \ images \ retrieval} \qquad (21)$$

$$R = \frac{Number \ of \ relevant \ images \ retrieved}{Total \ number \ of \ relevant \ images} \qquad (22)$$

To check the performance of our proposed technique, we use the average precision and recall. The quantitative measure defined is below

$$P(i) = \frac{1}{100} \sum_{1 \leq j \leq 1000, r(i,j) \leq 100, ID(j)=ID(i)} 1, \qquad (23)$$

where $P(i)$ is precision of query image $i$, $ID(i)$ and $ID(j)$ are class ID of image $i$ and $j$ respectively, which are in the range of 1-10. The $r(i,j)$ is the rank of image $j$ (it means position of image $j$ in the retrieved images for query image $i$, an integer between 1 and 1000). This value is percentile of images belonging to the category of image $i$, in the first 100 retrieved images. The average precision $P_t$ for category $t(1 \leq t \leq 10)$ is given by

$$P_t = \frac{1}{100} \sum_{1 \leq i \leq 1000, ID(i)=t} P(i) \qquad (24)$$

In order to evaluate the performance of our proposed method using the most relevant features set, the evaluation is carried out with each image in each class as query image and the number of retrieved images set as 20 to compute the precision and recall of each query image and finally obtain the average precision and average recall for 100 images per class. Comparison of average precision of the proposed method with other standard methods for Wang's dataset was shown in the Table 1. Our proposed method achieves better average precision and average recall values compare to methods shown in the table.

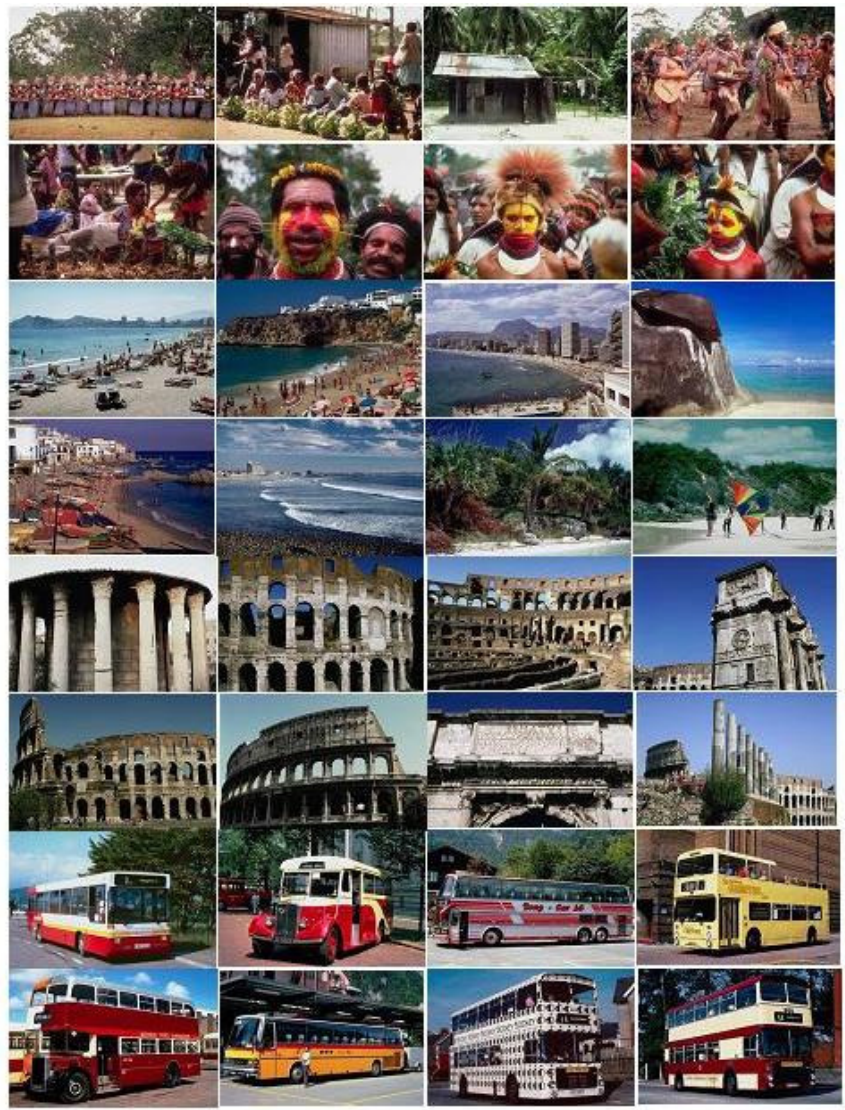

Figure 4. Sample images of Wang's dataset

The average precision of the retrieval results of the different images with number of returned images are plotted on the graph as shown Fig. 5. The precision value depends upon the total number of relevant images retrieved and hence it is directly proportional to number of relevant images retrieved images for a given query image. From the graph, it is observed that, our approach retrieves highest number of relevant images for a query image compared to other approaches. Hence, the precision value is high for the Wang's dataset using minimum distance classifier with Euclidean distance.

Table 1. Comparison of average precision with other standard methods for Wang's dataset

| Class | Image based SIFT-LBP [7] | Patch based SIFT-LBP [7] | Histogram Based [28] | Colour, shape, texture based [2] | Proposed Method |
|---|---|---|---|---|---|
| Africa | 0.57 | 0.54 | 0.30 | 0.48 | 0.56 |
| Beaches | 0.58 | 0.39 | 0.30 | 0.34 | 0.60 |
| Building | 0.43 | 0.45 | 0.25 | 0.36 | 0.58 |
| Bus | 0.93 | 0.80 | 0.26 | 0.61 | 0.94 |
| Dinosaur | 0.98 | 0.93 | 0.90 | 0.95 | 0.98 |
| Elephant | 0.58 | 0.30 | 0.36 | 0.48 | 0.66 |
| Flower | 0.83 | 0.79 | 0.40 | 0.61 | 0.88 |
| Horses | 0.68 | 0.54 | 0.38 | 0.74 | 0.78 |
| Mountain | 0.46 | 0.35 | 0.25 | 0.42 | 0.58 |
| Food | 0.53 | 0.52 | 0.20 | 0.48 | 0.54 |
| **Average** | **0.66** | **0.56** | **0.36** | **0.55** | **0.78** |

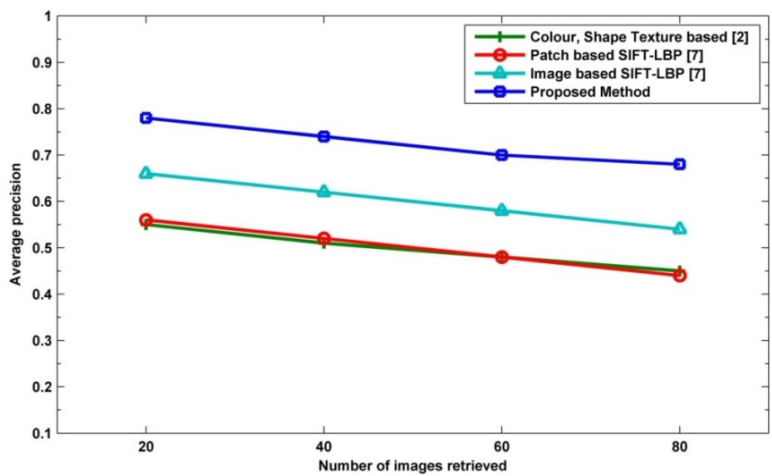

Figure 5. Comparison of average precision for proposed method with other methods

Similarly, the average recall of the retrieval results of the different images with number of returned images are plotted on the graph as shown Fig. 6. The recall value is directly proportional to number of relevant images retrieved out of total number of relevant images present in the dataset. It is observed that recall value of our approach is high compared to other approaches. This is because, our approach retrieves highest number of relevant images compared to other techniques.

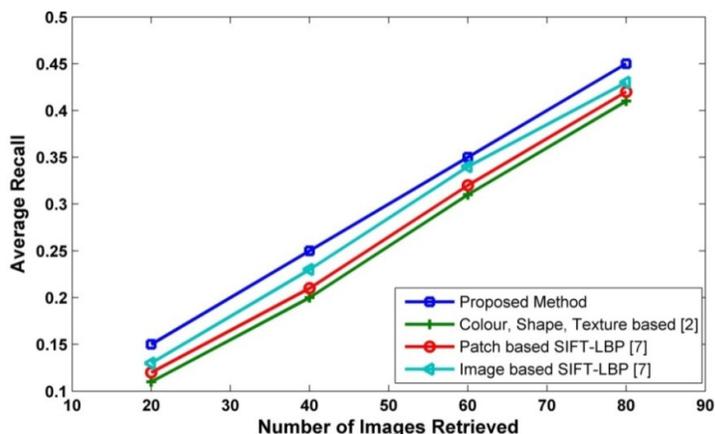

Figure 6. Comparison of average recall for proposed method with other methods

The confusion matrix calculated for Wang's dataset of Corel images. The confusion matrix shows the proportion in percentage, any class of image shown in a row is falsely detected as another class of image. It is observed that Flowers, Horses, Dinosaurs, Bus and Food images can be classified with high accuracy, while African peoples, Beach, Building, Elephant and Mountain images are easily confused with others. The average image retrieval rate of our proposed method is 84 % shown in Table 3 and the confusion matrix is shown in the Table 2.

Table 2. Confusion matrix for image retrieval in percentage from Corel dataset

| % Class | Africa % | Beach % | Building % | Bus % | Dinosaur % | Elephant % | Flower % | Horses % | Mountain % | Food % |
|---|---|---|---|---|---|---|---|---|---|---|
| Africa | 84 | 4 | 8 | | | | | | | 4 |
| Beach | 8 | 66 | 12 | 4 | | | | | 8 | 2 |
| Building | 8 | 6 | 76 | | | 4 | | | 4 | 2 |
| Bus | 6 | | 6 | 88 | | | | | | |
| Dinosaur | | | 2 | | 98 | | | | | |
| Elephant | | 2 | 8 | | | 86 | | 4 | | |
| Flower | | | | | | | 100 | | | |
| Horses | 2 | | 4 | | | | | 90 | | 4 |
| Mountain | | 14 | 12 | 4 | | 4 | | | 64 | 2 |
| Food | 6 | 2 | 2 | | | 2 | | | | 88 |

Table 3. Retrieval results of the proposed approach on Wnag's dataset Corel images with different distance measures

| Performance | Distance Measure | | | |
|---|---|---|---|---|
| | L1 | L2 | Canberra | Chi-square |
| Average Precession | 0.73 | 0.78 | 0.68 | 0.70 |
| Average retrieval rate (%) | 82 | 84 | 78 | 80 |
| Retrieval Time in Seconds /100 images | 10.4 | 10.1 | 10.8 | 10.5 |

## 5.2 Experiment on Caltech 256 database

The Caltech 256 dataset [30] contains 30,607 images are divided into 256 object classes. The images have high intra-class variability and high object location variability. Each category contains a minimum of 80 images and a maximum of 827 images. The mean number of images per category is 119. The images represent a diverse set of lighting conditions, poses, backgrounds and sizes. All the images are in JPEG format and in colour and average size of each image are 351x351 pixels, sample images are shown in Fig. 7.

In order to evaluate the performance of our proposed method using the most relevant features set, the evaluation is carried out with each image class as query image and the number of retrieved images set as 20 to compute the average retrieval rate for 100 images per class. We compared our method results with H-fusion [12], Color PHOW and Gray scale PHOW [33], [12] method and comparison results are shown in the Table 4. Our proposed method achieves better average retrieval rate compared to other methods shown in the Table 5. Note that the image retrieval rate for the Caltech 256 dataset is quite low, because this dataset has very high intra class variability and in several cases the object occupies a small portion of the image.

Table 4. Comparison of average retrieval rate (%) with other methods on Caltech 256 dataset

| Techniques | Average Retrieval Rate (%) |
|---|---|
| H-fusion | 33.6 |
| Color PHOW | 29.9 |
| Gray scale PHOW | 25.9 |
| Our Proposed Method | 42.6 |

Table 5. Retrieval results on Caltech 256 dataset with different distance measures

| Performance | Distance Measure | | | |
|---|---|---|---|---|
| | L1 | L2 | Canberra | Chi-square |
| Average retrieval rate (%) | 40.8 | 42.6 | 38.5 | 40.2 |
| Retrieval Time in Seconds /100 images | 14.5 | 13.6 | 15.2 | 15.6 |

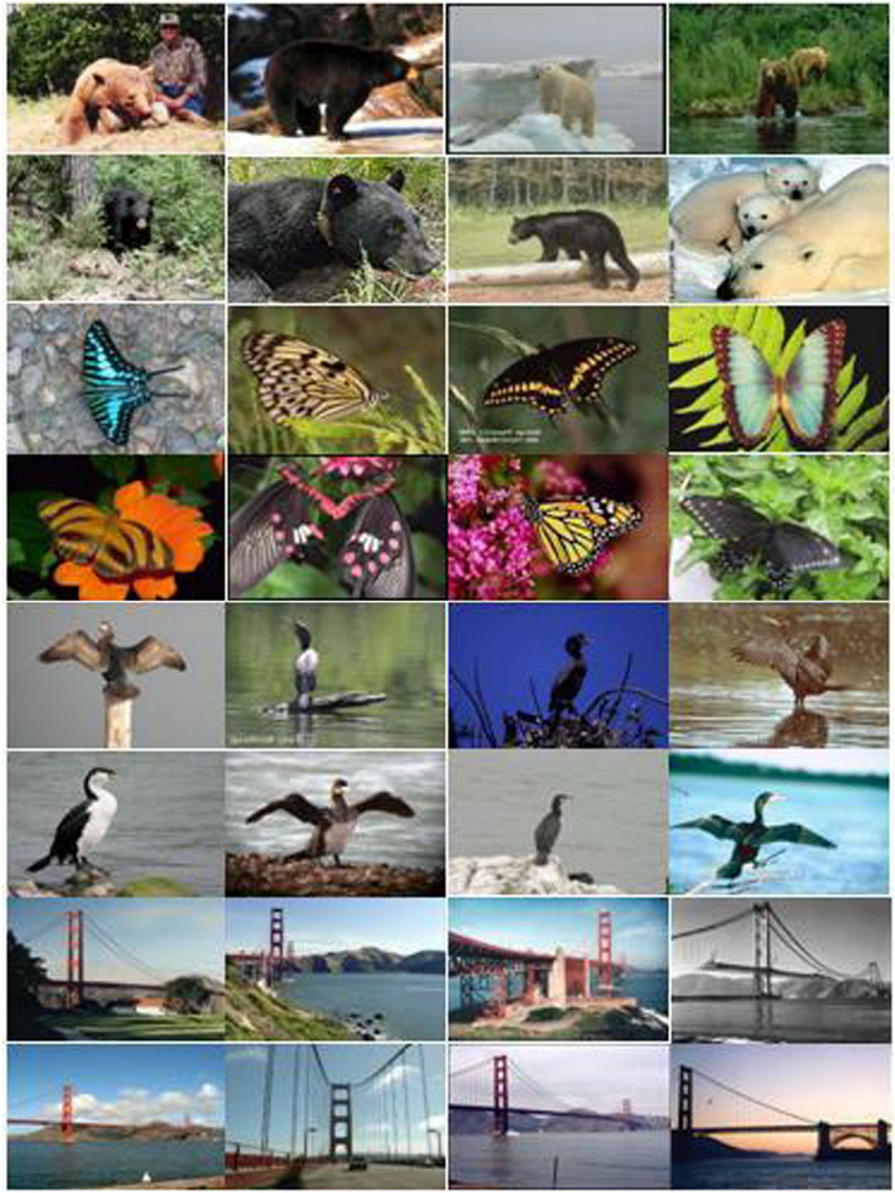

Figure 7. Sample database of Caltech 256 dataset

## 6. CONCLUSION

In this paper, we proposed a novel approach for content based image retrieval based on low-level features such as colour texture and shape. The DBC, Haar wavelet and HOG techniques are employed sequentially in order to extract colour texture and shape features from the image. The experiments are conducted using two benchmark databases such as Wang's and Caltech 256. The performance of our approach on Wang's dataset is evaluated using precision, recall, retrieval rate and processing time based on average results obtained for all the experiments. Similarly, Caltech 256 is used for experiments in order to verify the effectiveness of our approach based on comparison with other methods using retrieval rate. The evaluation results supports to claim that in all the experiments, our approach outperforms other approaches for image retrieval. This is due to the fact that the DBC captures the spatial relationship between any pair of neighbourhood pixels in a local region along a given direction and also it can extract more edge information than LBP. Further, the HOG descriptor stores the information about the shapes contained in the image. The combination of low-level features such as texture and shape provide accurate representation of content of an image, which helps to achieve high retrieval rate compared to other existing methods for standard datasets.

## ACKNOWLEDGEMENTS

The authors would like to thank the reviewers for their valuable comments and suggestions which helped lot to improve previous version of the paper.